\documentclass[letterpaper, 10 pt, conference]{ieeeconf}
\pdfoutput=1
\IEEEoverridecommandlockouts                              
\usepackage{gensymb}
\usepackage{cclicenses, graphicx}
\newcommand{\graspit}{\textit{GraspIt!}}
\newcommand{\mystep}[1]{{\vspace{2mm}\noindent\textbf{#1}}}

\title{\LARGE \bf On the Feasibility of Wearable Exotendon Networks
  for Whole-Hand Movement Patterns in Stroke Patients}

\author{Sangwoo Park$^{1}$, Lauri Bishop$^{2}$, Tara Post$^{2}$, Yuchen Xiao$^{1}$, Joel Stein$^{2,3}$ and Matei Ciocarlie$^{1,3}$%
\thanks{*This work was supported in part by the Columbia-Coulter Translational
Research Partnership and by the National Science Foundation grant number IIS-1526960.}%
\thanks{$^{1}$Department of Mechanical Engineering, Columbia University, New York, NY 10027, USA.}%
\thanks{\hspace{-3mm}{\tt\small \{sp3287, yx2281, matei.ciocarlie\}@columbia.edu}}%
\thanks{$^{2}$Department of Rehabilitation and Regenerative Medicine, Columbia University, New York, NY 10027, USA. {\tt\small \{tp2213, lb2413, js1165\}@cumc.columbia.edu}}%
\thanks{$^{3}$Co-Principal Investigators}
}

\begin{document}

\maketitle
\thispagestyle{empty}
\pagestyle{empty}

\begin{abstract}
Fully wearable hand rehabilitation and assistive devices could extend
training and improve quality of life for patients affected by hand
impairments. However, such devices must deliver meaningful
manipulation capabilities in a small and lightweight package. In this
context, this paper investigates the capability of single-actuator
devices to assist whole-hand movement patterns through a network of
exotendons. Our prototypes combine a single linear actuator (mounted
on a forearm splint) with a network of exotendons (routed on the
surface of a soft glove). We investigated two possible tendon network
configurations: one that produces full finger extension (overcoming
flexor spasticity), and one that combines proximal flexion with distal
extension at each finger. In experiments with stroke survivors, we
measured the force levels needed to overcome various levels of
spasticity and open the hand for grasping using the first of these
configurations, and qualitatively demonstrated the ability to execute
fingertip grasps using the second. Our results support the feasibility
of developing future wearable devices able to assist a range of
manipulation tasks.

\end{abstract}

\section{INTRODUCTION}

Wearable devices have established themselves as an important area of
focus for research in robotic rehabilitation. Traditional
robot-assisted therapy, based on desktop-sized (or larger) machines,
can mainly be provided in clinical settings. In contrast, wearable
devices promise to enable use outside the hospital, providing the
larger number of repetitions that is generally considered key to
effective rehabilitation~\cite{LANG07}. Beyond rehabilitation,
wearable devices could also act as functional orthoses, providing
assistance with Activities of Daily Living and increasing
independence.

In this paper, we focus on wearable assistive devices for the
hand. This is a particularly challenging context: the human hand is
highly dexterous, often modeled as having as many as 20 individual
joints. This high dimensionality of the joint position space gives
rise to an enormous set of possible configurations.

A key tenet of our approach is that a hand orthosis can provide
meaningful assistance with daily manipulation tasks even when using a
number of actuators far smaller than the number of joints in the
hand. Even though the human hand is highly dexterous, previous
research suggests that numerous manipulation tasks are dominated by a
smaller number of effective degrees of
freedom~\cite{SANTELLO98,TODOROV04,THAKUR08}. Previous work has shown
that this result translates to artificial hands as
well~\cite{CIOCARLIE09}, where is it often implemented using the key
principles of underactuation and passive compliance. In recent work,
In et al.~\cite{hyunki2015b} have shown how underactuation can also be
applied to our area of interest, namely tendon-driven assistive
devices for the human hand. Overall, it seems likely that using a
relatively small number of motors will be key in achieving a compact
and lightweight wearable device.

\begin{figure}[t]
\centering
\setlength{\tabcolsep}{0mm}
\begin{tabular}{c}
\begin{tabular}{ccc}
\includegraphics[width=0.33\linewidth]{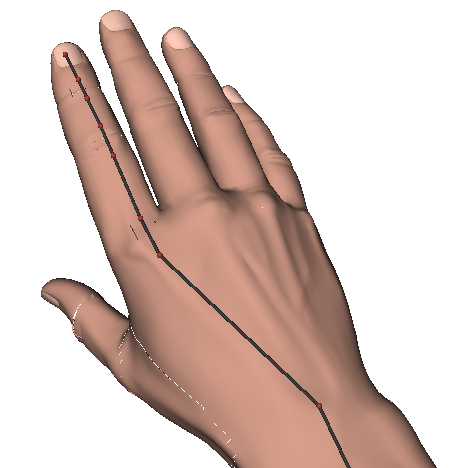}&
\includegraphics[width=0.33\linewidth]{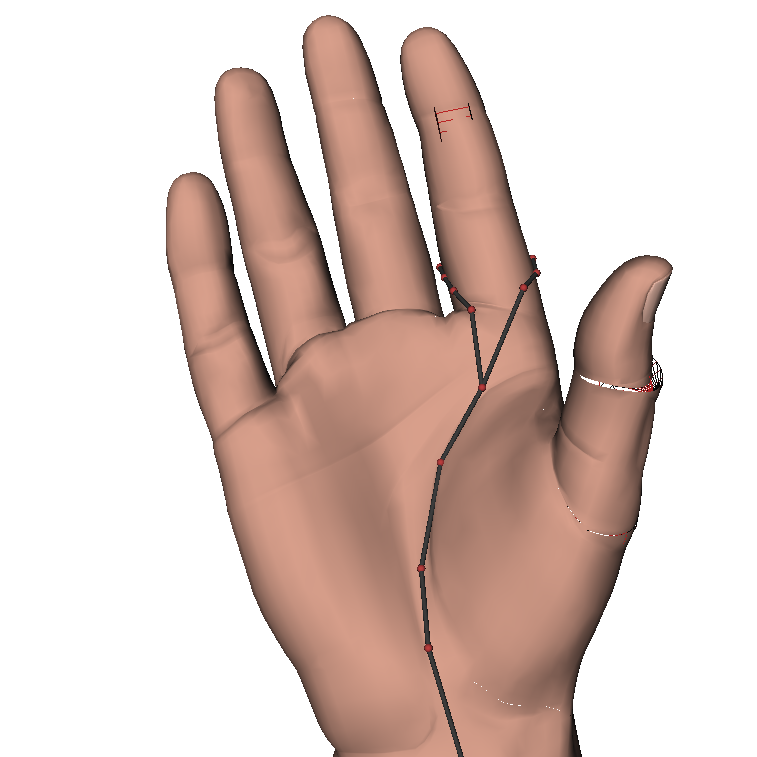}&
\includegraphics[width=0.33\linewidth]{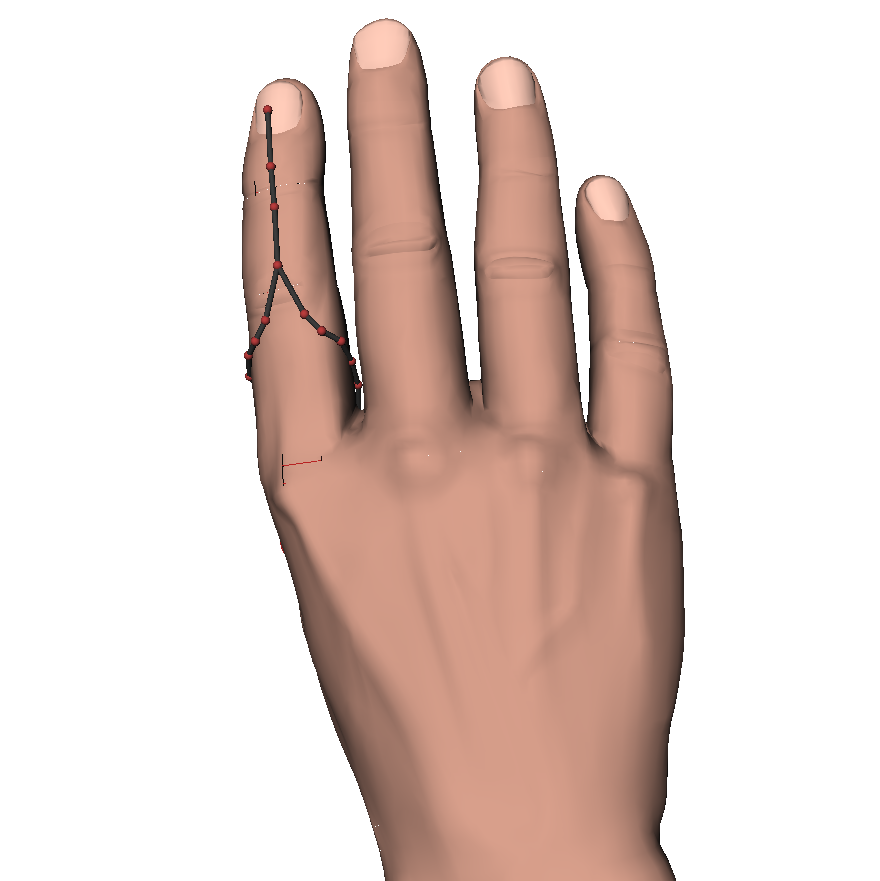}
\end{tabular}\\
\includegraphics[width=1.0\linewidth]{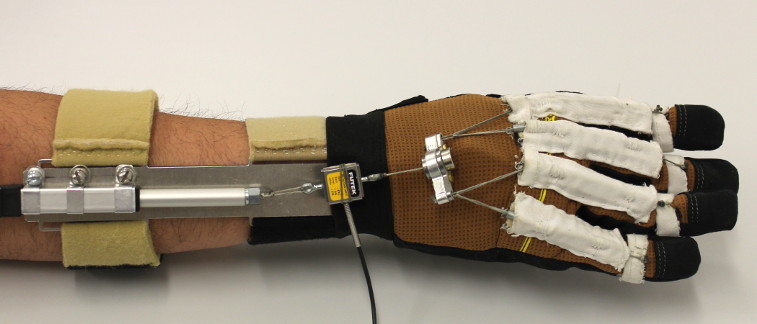}
\end{tabular}
\caption{Example exotendon routes for achieving different movement
  patterns illustrated for index finger (top) and whole-hand movement
  pattern implemented with single-actuator exotendon network
  (bottom). }
\label{fig:candy}
\end{figure}

To implement these principles, we used a network of exotendons, or
tendons routed on the surface of the hand, used to initiate and assist
movement. The tendons form a network providing both intra- and
inter-finger underactuation; a subset of them are connected to
actuators mounted on the forearm (Fig.~\ref{fig:candy}). However,
before such devices become practical, key questions still need to be
addressed. \textit{First, can a device using few and relatively small
  motors reach the force levels needed for meaningful assistance?}
This is a particularly important question given that a common stroke
aftereffect is hand spasticity, with permanent involuntary
flexion. \textit{Second, can we hope to achieve the dexterity levels
  needed to enable varied and useful manipulation}, across a wide
range of patients exhibiting different impairment patterns?

In the study presented here, we implemented and tested devices that
assist with two movement patterns (full hand extension and fingertip
pinch). These aim to address some of the hand impairment types most
commonly encountered as stroke aftereffects. Each pattern assists with
movement of multiple fingers (four and five, respectively) through the
use of a single actuator. Overall, the main contributions include the
following:
\begin{itemize}
\item We investigate the feasibility of providing assistance for
  whole-hand movement patterns using a single actuator and a network
  of exotendons, and report here on prototype designs that achieve this
  functionality.
\item We quantitatively assess the combined actuation force needed
  for assisting a multi-digit hand movement pattern (hand extension)
  in stroke patients. It is, to the best of our knowledge, the first
  time that exotendon force needed to overcome spasticity has been
  measured and reported. We believe this data will prove highly
  significant as we make progress towards dexterous, yet compact and
  wearable assistive devices for the hand.
\item We correlate our results with the spasticity level observed in
  stroke patients, measured using the Modified Ashworth Scale,
  commonly employed in patient assessment for rehabilitation. This
  type of data will help identify patient populations most suited for
  using wearable assistive devices for the hand. We also characterize
  the resistance to movement provided by spastic muscles through the
  assisted range of motion, further informing future designs.
\item We provide qualitative results indicating the feasibility of
  single-actuator assistance for a second movement pattern (fingertip
  pinch).
\end{itemize}

\section{RELATED WORK}

Wearable assistive devices for the hand have been proposed in the
literature using various actuation methods (e.g. electric, pneumatic,
etc.) and transmission mechanisms (e.g. linkages, tendons,
etc.). While linkage driven systems exhibit efficient power
transmission and support bidirectional actuation, this type of devices
often faces the additional challenge of rotational axes
misalignment. Several methods have been proposed to address this, such
as direct matching of joint centers in HANDEXOS~\cite{chiri2012} and
HEXORR~\cite{schabowsky2010}, remote center of motion mechanism in
EHI~\cite{fontana2009}, and serial links chain connected to distal
phalanges in HEXOSYS~\cite{iqbal2014}. While effective for
rehabilitation exercises, complex linkages also increase size and thus
reduce applicability in constrained, cluttered environments typical of
daily living. We note that wearable linkages can also take the form of
a supernumerary robotic finger~\cite{irfan2015}, a different way of
providing assistance without interfering with the natural kinematics
of the hand.

In contrast, wearable hand orthoses comprising only soft structures
produce more compact systems, since soft devices do not require
appropriate alignment with the biological joints of the
user~\cite{hyunki2015a}. Fabrics can provide robust structure while
keeping the device adjustable and
affordable~\cite{polygerinos2015}. Easy customization is important, as
it allows one device to be used with multiple users.

Hand rehabilitation devices using soft pneumatic
actuators~\cite{polygerinos2015, hong2015} keep the advantages of
completely soft, wearable robots, but pneumatic actuators require an
additional air compressor. Our approach combines soft gloves with
guided tendons driven by linear electric actuators. The tendon-driven
design allows us to mount actuators at locations proximal to the
joints they are driving, thus making the device easier to wear. A
tendon-based approach also simplifies the construction of
underactuated kinematics, as tendons can cross multiple joints for
intra-finger coupling and bifurcate for inter-finger coupling. A
number of existing hand and arm rehabilitation devices are also tendon
driven~\cite{jones2014, baker2011, ying2012, delph2013}. However,
aiming towards fully wearable solutions, our device is fully
self-contained and includes lightweight, portable actuators.

A research effort closely related to our approach resulted in the
development of the BiomHED prototype~\cite{sangwook2014}. This
exotendon-driven device uses 7 motors attached to a tendon network
that mimics the geometry of hand muscle-tendon units. Experiments
established the ability of the device to generate fingertip motion and
increase finger workspace in stroke survivors, highly encouraging for
the area of active hand orthoses. However, the BiomHED prototype used
one actuator per finger, and implemented a single movement
pattern. Here, we investigate multiple whole-hand movement patterns
each driven by a single actuator, aiming to reduce the number of
actuators needed by future, more dexterous devices.

In terms of quantitative assessment of actuation forces for hand
movemebt, Iqbal et al.~\cite{iqbal2014} conducted a series of
experiments to measure the forces applied on each finger by a hand
exoskeleton device, showing a maximum force of up to 45N. In et
al.~\cite{hyunki2013} showed that a soft robot hand with joint-less
structure can apply effective joint forces for grasping. Recently,
pinch and enveloping grasp force by a soft hand orthosis with a tendon
routing system, the Exo-Glove~\cite{hyunki2015b}, were measured on a
healthy subject, with EMG signals confirming that the subject did not
exert additional voluntary force. Our results are based on studies
with stroke survivors, and are additionally correlated with spasticity
levels, which can significantly affect manipulation
abilities. Finally, a comprehensive review of additional hand
exoskeletons for rehabilitation and assistance can be found in the
study by Heo et al.~\cite{pilwon2012}.

\section{MOVEMENT PATTERNS AND TENDON NETWORKS}

Stroke survivors experience a broad range of hand impairments, ranging
from barely perceptible slowing of fine finger movements to complete
loss of all voluntary movement. It is unlikely, for now, that a single
device can effectively address all these impairment types. We have
thus chosen to focus on certain patterns of impairments that are
particularly common and challenging from a rehabilitation perspective,
selected based on the clinical experience of our team. We describe
these below, noting again that there are a wide range of other motor
and functional impairments that affect stroke survivors.

\begin{itemize}
\item Pattern A: These individuals are able to form a gross grasp with
  the hand moving all digits in synergy, but lack sufficient finger
  extension to actively open the hand after grasping. This pattern is
  particularly challenging as it commonly includes spasticity, where
  the hand is in a persistently flexed pose and digit extensors are
  unable to overcome ongoing involuntary contraction of the
  flexors. These individuals also typically lack individuated finger
  movements.
\item Pattern B: Another common pattern is the ability to move all
  digits of the hand, but to have limited individuation, and for the
  movements to be slow, lacking in dexterity, and of diminished
  force. Such an individual may be able to oppose the thumb to each of
  the other digits in sequence, but only slowly and with considerable
  effort. The ability to manipulate objects is limited.
\end{itemize}

\begin{figure}[t]
\centering
\begin{tabular}{cc}
\raisebox{-17mm}{\includegraphics[width=40mm]{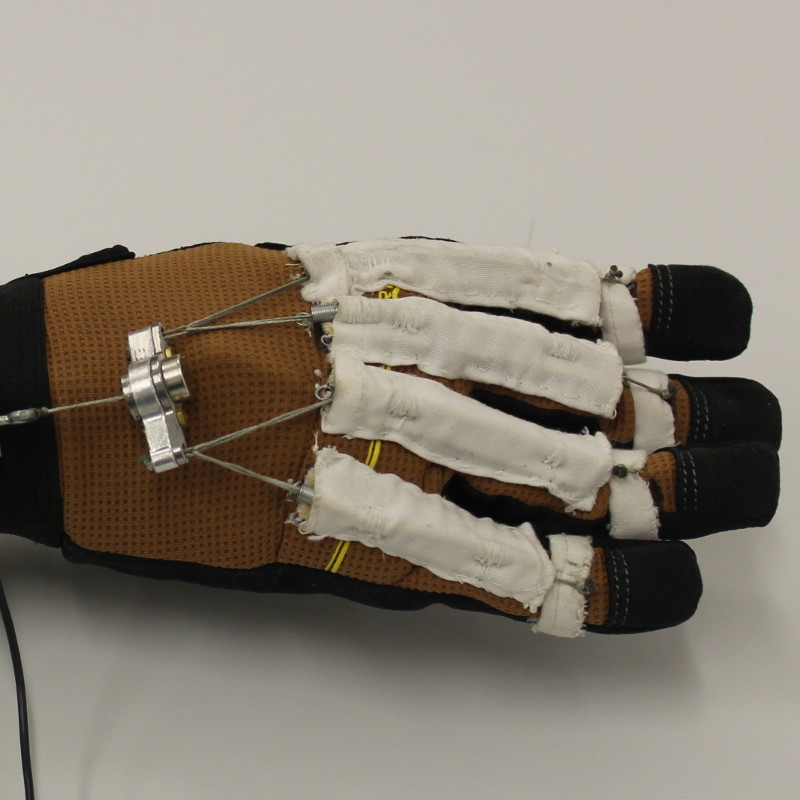}}
\begin{tabular}{c}
\includegraphics[width=40mm]{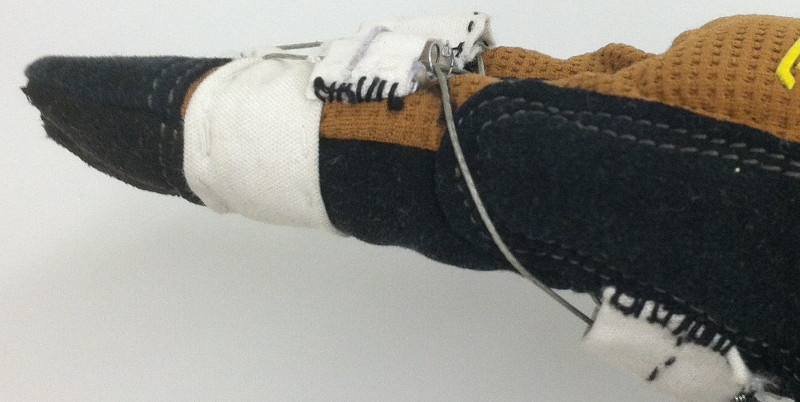}\\
\includegraphics[width=40mm]{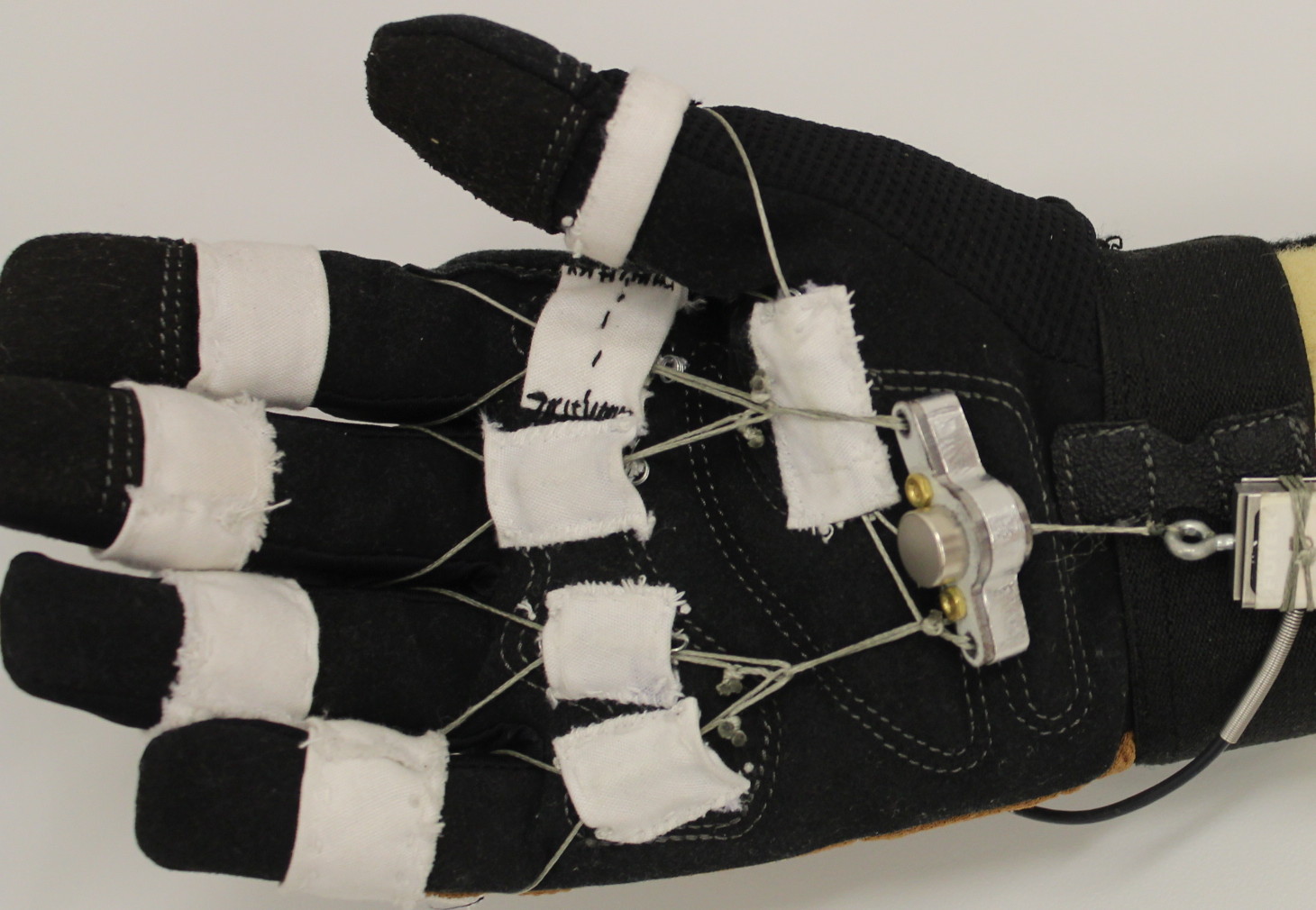}
\end{tabular}
\end{tabular}
\caption{Hand exotendon configurations for eliciting desired movement
  patterns. Left: tendon configuration 1 (hand extension), dorsal
  view. Right: tendon configuration 2 (MCP flexion / IP extension), lateral
  and palmar views.}
\label{fig:config}
\end{figure}

Here, we report on two exotendon network configurations informed by
these patterns. Each of these configurations is designed to be driven
by a single actuator, with multiple joints moving in synergy. For
initial study and assessment, we have implemented each configuration
separately, in a dedicated prototype. However, combined versions able
to produce multiple movement patterns with few actuators will be a
promising direction for future research. Both configurations are
illustrated in Fig.~\ref{fig:config}.

\mystep{Tendon configuration 1}: hand extension. In this
configuration, one motor assists extension for all digits. From a
clinical perspective, this configuration addresses Pattern A described
earlier, where a person lacks sufficient finger extension to overcome
spasticity and actively open the hand. Combined finger extension is
amenable to direct implementation through a single motor, since a
tendon can be routed on the dorsal side of all joints all the way to
the phalanges. These routes also allow the tendons to be neutral with
regard to abduction/adduction motion of the fingers. Simulations
carried out using the human hand model included with the \graspit{}
simulator for robotic grasping~\cite{MILLER04} have shown that
complete range of motion of all the joints of the index finger
requires 57mm of travel of the tendon. This matches the specifications
of the off-the-shelf linear actuators we use, noting that functional
use of the hand for common tasks is unlikely to require full
simultaneous extension of all joints.

The implementation we report on in this study addresses all digits
except for the thumb. The trapeziometacarpal joint is significantly
more complex than finger carpometacarpal joints; \graspit{}
simulations based on the common model with two non-perpendicular axes
of rotation indicate that most extensor exotendon routes will also
have a limited but non-zero effect on thumb abduction. We are
currently investigating this effect and plan to include the thumb in
future prototypes developed for functional testing.

\mystep{Tendon configuration 2}: metacarpophalangeal (MCP) flexion /
interphalangeal (IP) extension. This pattern assumes opposite motion
at the MCP joints versus the IP joints for each finger. Functionally,
this pattern can allow transition between enveloping postures and
fingertip opposition postures. For stroke patients exhibiting pattern
B described earlier, this could increase the range of grasps that can
be executed. From an implementation perspective, it is achieved
through a single tendon for each finger, routed on the palmar side of
the MCP joint then wrapping around the finger to the dorsal side of
the proximal and distal IP joints. The tendon bifurcates to wrap
around both sides of the finger in order to obtain a neutral effect on
finger adduction. Our implementation of this pattern addresses all
digits, including the thumb, where the role of MCP flexion is instead
played by adduction.

\section{PROTOTYPE DESIGN AND FABRICATION}

\begin{figure}[t]
\centering
\begin{tabular}{r}
\includegraphics[width=0.945\linewidth]{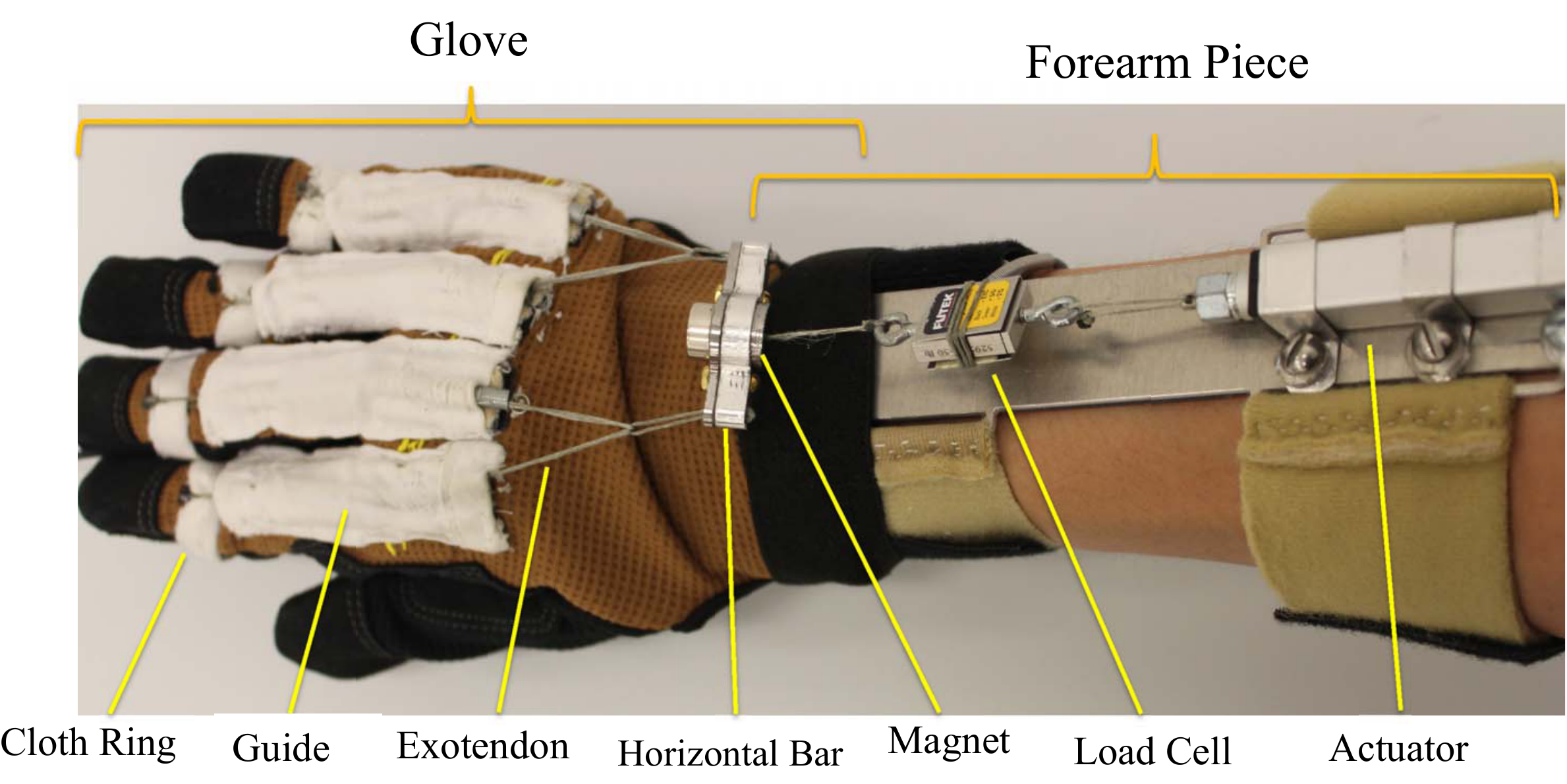}\\
\includegraphics[width=0.9\linewidth]{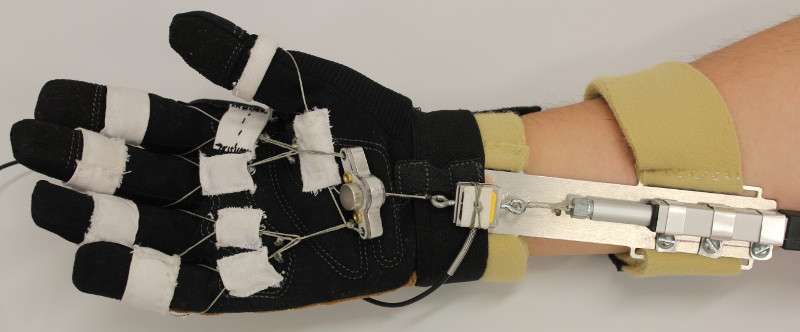}
\end{tabular}
\caption{Prototype hand orthotic devices. Top: tendon configuration
  1. Bottom: tendon configuration 2. Both devices comprise a forearm
  splint with a mounted actuator and a glove implementing the desired
  tendon network.}
\label{fig:module}
\vspace{-2mm}
\end{figure}

Our overall design is illustrated in Fig.~\ref{fig:module}. To
facilitate donning the device, we split it into two modules: a forearm
piece with actuation, and a glove with the tendon network. The two
components are connected via mechanical features that automatically
detach before potentially dangerous forces are reached. This
mechanical coupling includes a permanent magnet connecting the motor
and exotendons. We currently use permanent magnets capable of a pull
force of either 34N (D73, KJ Magnetics Inc.)  or 41N (D73-N52, KJ
Magnetics Inc.), both cylindrical with a diameter of 11 mm and
thickness of 4.7 mm. Connector pieces with different tendon lenghts
allow us to adjust the device to the subject, such that, with all
digits fully flexed and the actuator in the fully extended position,
we remove all tendon slack up to a few millimeters.

\begin{figure}[t]
\centering
\begin{tabular}{c}
\includegraphics[width=0.8\linewidth]{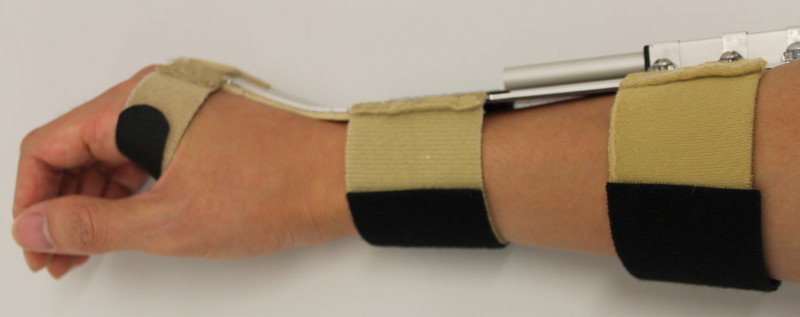}\\
\includegraphics[width=0.8\linewidth]{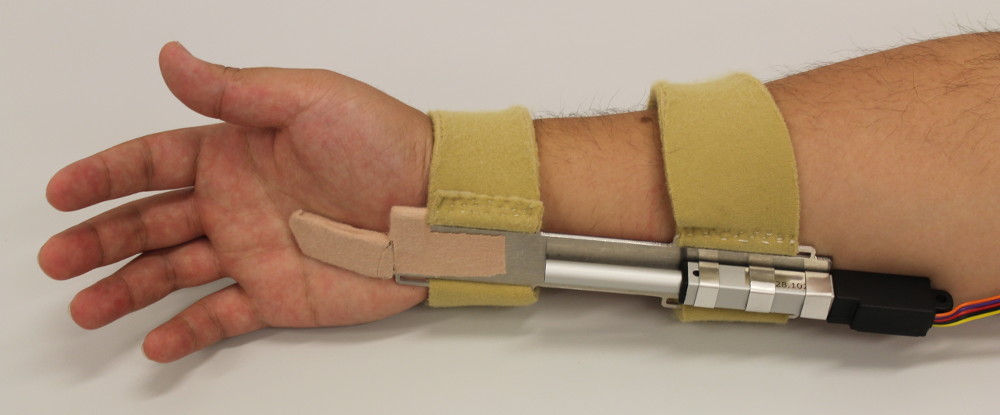}
\end{tabular}
\caption{Forearm splint components for devices with tendon
  configuration 1 (top) and tendon configuration 2 (bottom).}
\label{fig:forearm}
\end{figure}

A linear actuator with a 50 mm stroke length, a 5 mm/s maximum speed
of travel, and a 50 N peak force (Firgelli, L12-P-50-210-12) is
mounted on the forearm piece. A 50mm stroke length suffices for the
expected range of motion, and 5mm/s as maximum speed is slow enough to
prevent any hazardous circumstances. The peak force of 50N is above
that of the breakaway magnetic coupling, so it was never reached in
our experiments.

One of the main roles of the forearm piece is to constrain the wrist
joint~(Fig \ref{fig:forearm}). Splinting the wrist is important in our
mechanism as it helps extend the fingers without hyperextending the
wrist. Furthermore, the splint is designed to maintain a wrist
extension angle of 30\degree, considered a functional wrist
pose~\cite{natasha2003}. This design also reduces distal migration,
or the phenomenon where an entire orthotic device slowly slides
towards the distal end of the arm while in use. To reduce pressure,
which might cause pain on the hand, soft materials, such as moleskin,
are attached underneath the splint.

\begin{figure}[t]
\centering
\begin{tabular}{cc}
\includegraphics[height=31mm]{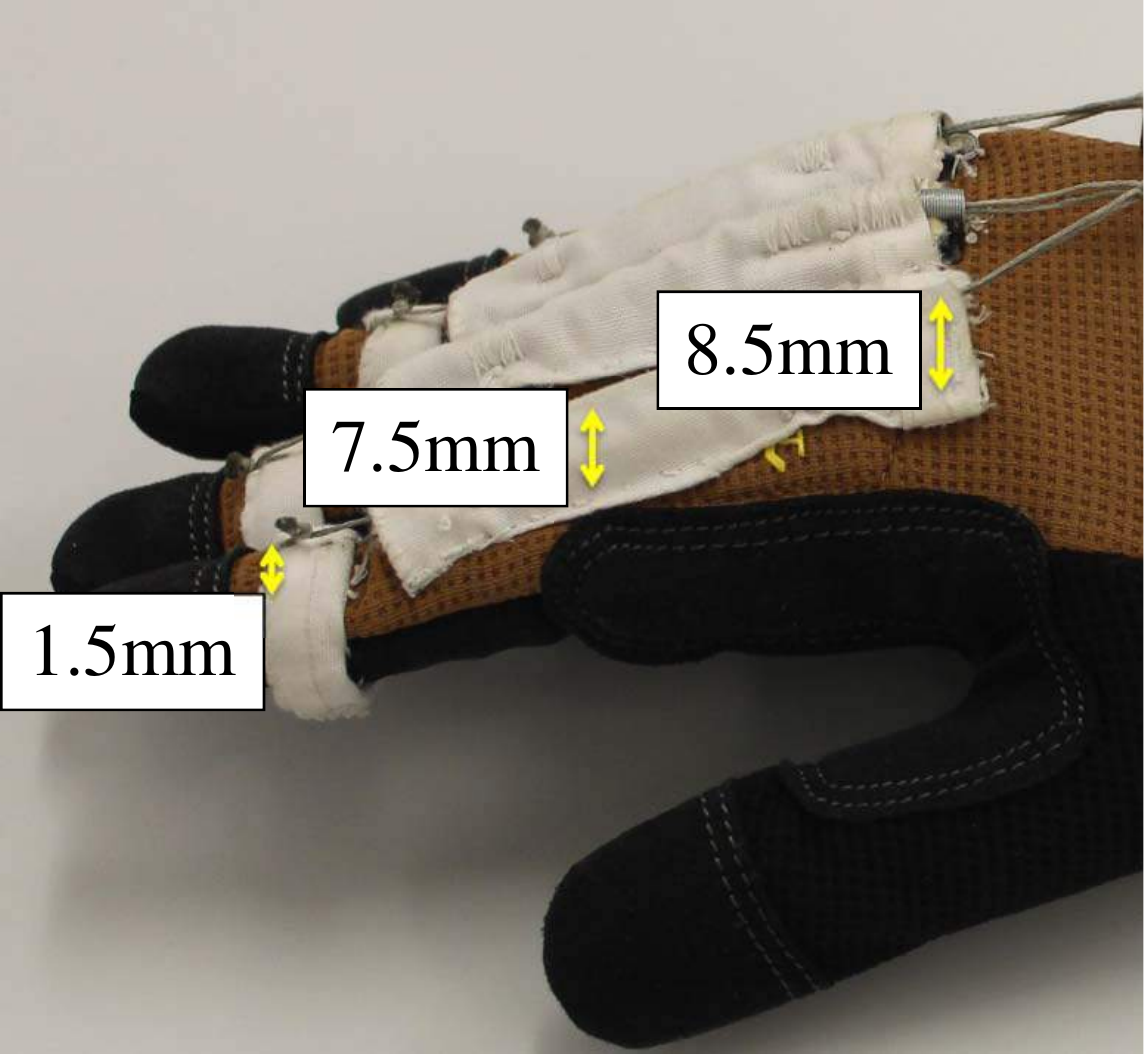}
\includegraphics[height=31mm]{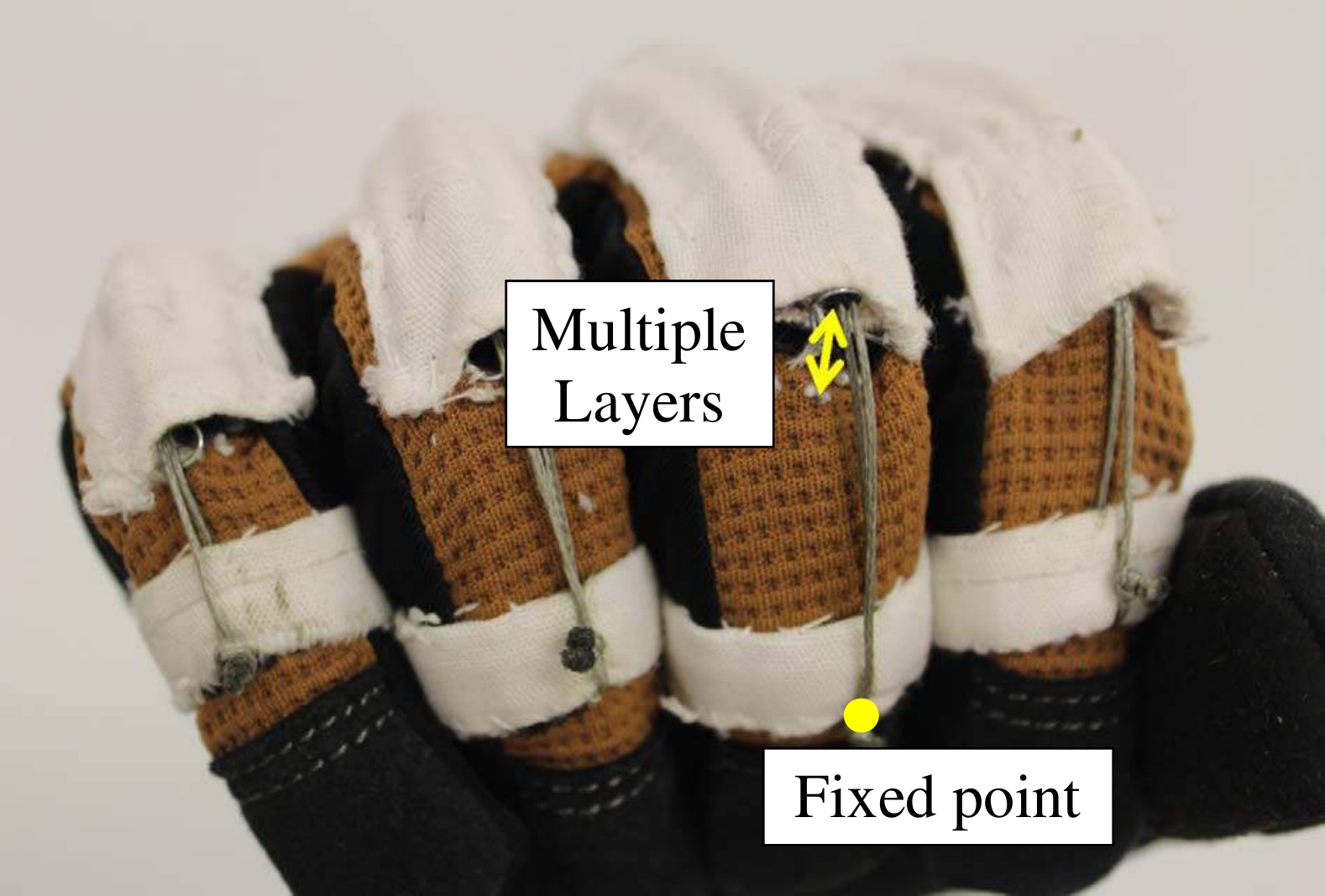}
\end{tabular}
\caption{Tendon guides with increased moment arms via raised
  pathways.}
\label{fig:glove}
\end{figure}

The tendon networks described in the previous section are implemented
on the glove component of the device. In each case, one tendon
connected to the actuator bifurcates into a network that actuates each
finger. All bifurcation points are rigid, with no differential
mechanism installed to distribute loads. A number of existing
cable-driven soft wearable hand devices have tendon attachment points
on the fingertip of the glove~\cite{hyunki2015a,delph2013}. We have
found that this design can produce finger hyperextension at the DIP
joint. To alleviate this problem, the tendons attach on each finger to
a cloth ring on the middle phalanx. Through IP joint coupling, this
produces both PIP and DIP extention, without causing hyperextension.

On the dorsal side of each finger, raised tendon guides sitting on top
of multiple layers of fabric are used to increase the moment arm of
the extensor tendons around the joints. The increased moment arms
allow us to reduce the linear forces applied to the tendons. For the
index finger (and representative for the other fingers), the raised
pathways have height of 8.5mm above the MCP joint and 7.5mm above the
PIP joint; the cloth ring only protrudes 1.5mm above skin (Fig
\ref{fig:glove}). For tendons on the palmar side of the joints, we
have found that such increased moment arms are not necessary.

A load cell (Futek, FSH00097) is installed between the actuator and
the magnet piece to measure the tension of the actuated tendon. The
sensor has been calibrated to have a resolution of 0.196N and can
measure up to 50N.

\section{EXPERIMENT DESIGN}

Our experiments were designed to provide initial validation of the
approach with the intended target population of stroke patients. In
particular, we aimed to verify the capability of the device to produce
the expected patterns and ranges of motion, and to characterize the
forces encountered, especially when assisting patients exhibiting
various levels of spasticity.

Testing was performed with five stroke survivors, three female and two
male. All testing was approved by the Columbia University Internal
Review Board, and performed in a clinical setting under the
supervision of Physical and/or Occupational Therapists. All subjects
displayed right side hemiparesis following a stroke event; in all
cases, experiments took place more than 6 months after the
stroke. Subjects also exhibited different spasticity levels, ranging
between 1 and 3 on the Modified Ashworth Scale (MAS).

The first step in the experimental procedure consisted in measuring
the patient's range of motion in all digits as well as the wrist, and
assessing the spasticity level on the MAS. The next step consisted of
donning the orthotic device, consisting of the forearm splint and the
exotendon glove. After donning, the motor on the forearm splint was
connected to the tendon network via a breakaway magnetic mechanism as
described earlier.

Starting with the linear actuator at full extension, we define one
trial as one excursion of the actuator to the completely retracted
position. Depending on the tendon network being used (tendon
configurations 1 or 2 described above), this produced a given movement
pattern of the subject's hand. Throughout each trial, we recorded both
the actuator position and the tendon force levels reported by the load
cell; both measurements were taken at a frequency of 100 Hz. When
tendon forces exceeded the maximum load supported by the magnet, the
breakaway mechanism disengaged and the actuator retraction completed
without exerting any forces to the subject.

\begin{figure}[t]
\centering
\begin{tabular}{c}
\includegraphics[width=0.8\linewidth]{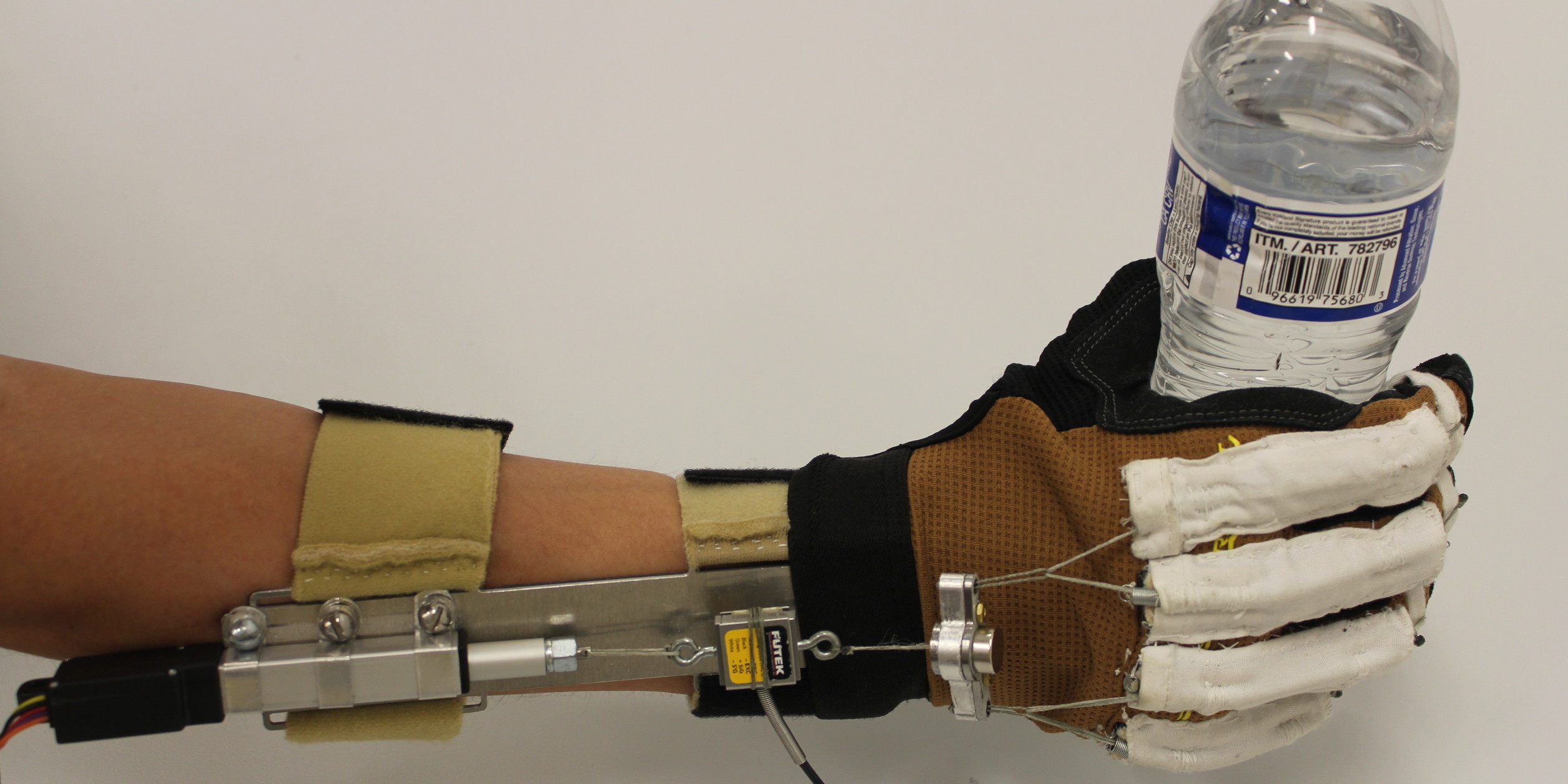}\\
\includegraphics[width=0.8\linewidth]{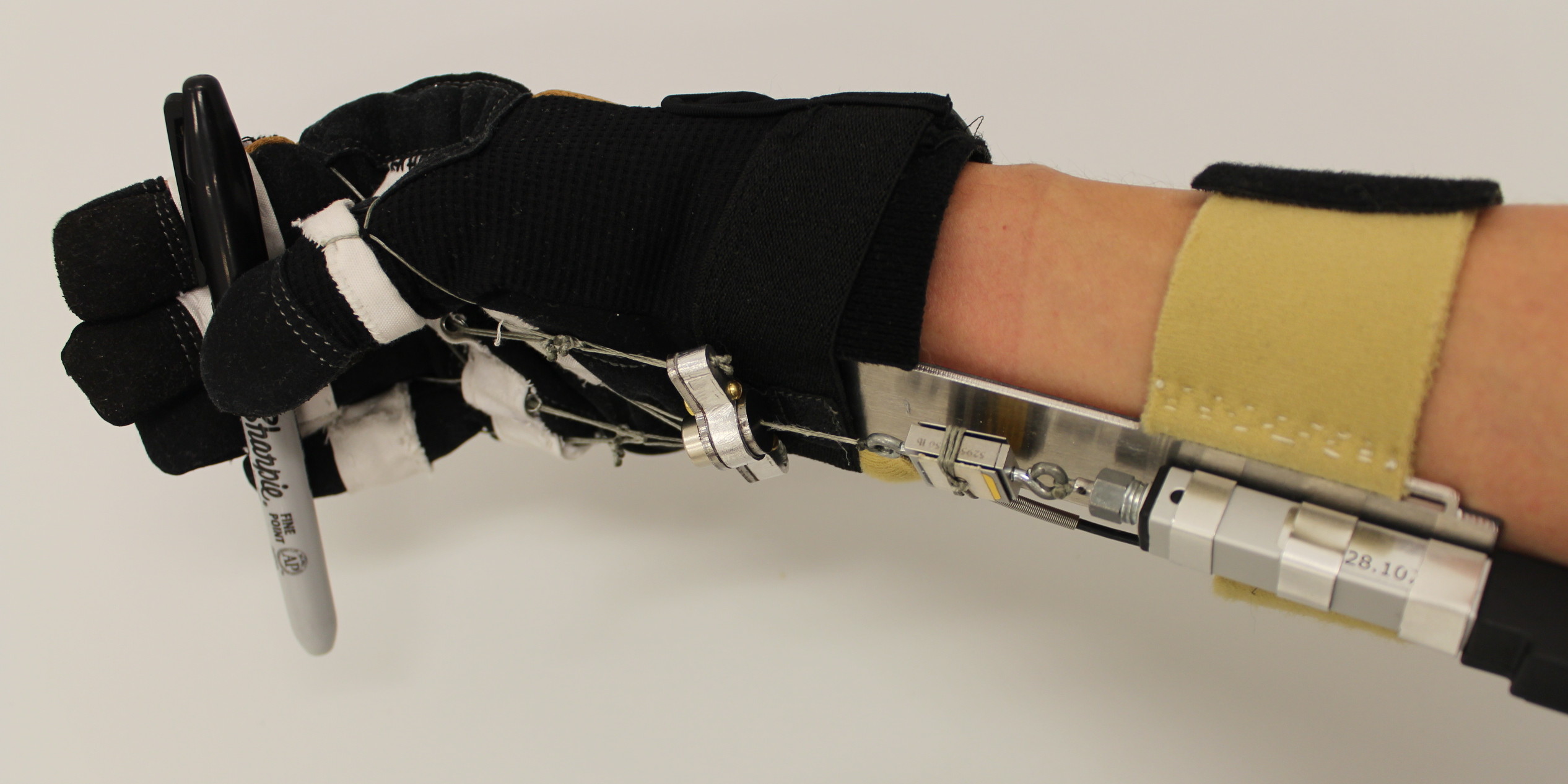}
\end{tabular}
\caption{Illustrations of functional experiments with tendon
  configuration 1 (top) and tendon configuration 2 (bottom).}
\label{fig:exper}
\end{figure}

With each subject, we performed the following set of trials:
\begin{itemize}
\item 1 trial where we asked the subject to relax their hand and not
  apply any voluntary forces;
\item 2-3 trials where we asked the subject to voluntarily assist the
  device in producing the intended movement pattern, to the best of
  their abilities;
\item (for Configuration 1) 2-3 trials where the subject attempted to
  grasp an object (soda can). Starting from the subject's rest pose,
  the exotendon was engaged by retracting the linear actuator,
  providing finger extension. Once functional extension was achieved,
  the hand was positioned around the object and the exotendon was
  released allowing the subject to flex the fingers (illustrated by an
  able-bodied user in Fig.~\ref{fig:exper}). If needed, the subject
  was assisted by the experimenter in positioning the arm such that
  the hand would be able to execute the grasp.
\item (for Configuration 2) 2-3 trials where the subject attempted to
  execute a pinch grasp of an object (highlighter pen). Starting from
  the subject's rest pose, the exotendon was engaged by retracting the
  linear actuator, placing the hand in a pose appropriate for
  fingertip grasping a given object (illustrated by an able-bodied in
  Fig.~\ref{fig:exper}). If needed, the object was positioned by the
  experimenter such that the hand would be able to execute the grasp.
\end{itemize}

After the completion of the procedure, subjects were asked to describe
their impressions of wearing the device, any discomfort or pain
produced by it, and any suggestions for improvement.

\section{RESULTS}

\begin{figure*}[t]
\setlength{\tabcolsep}{-2mm}
\centering
\begin{tabular}{ccc}
\includegraphics[width=0.35\linewidth]{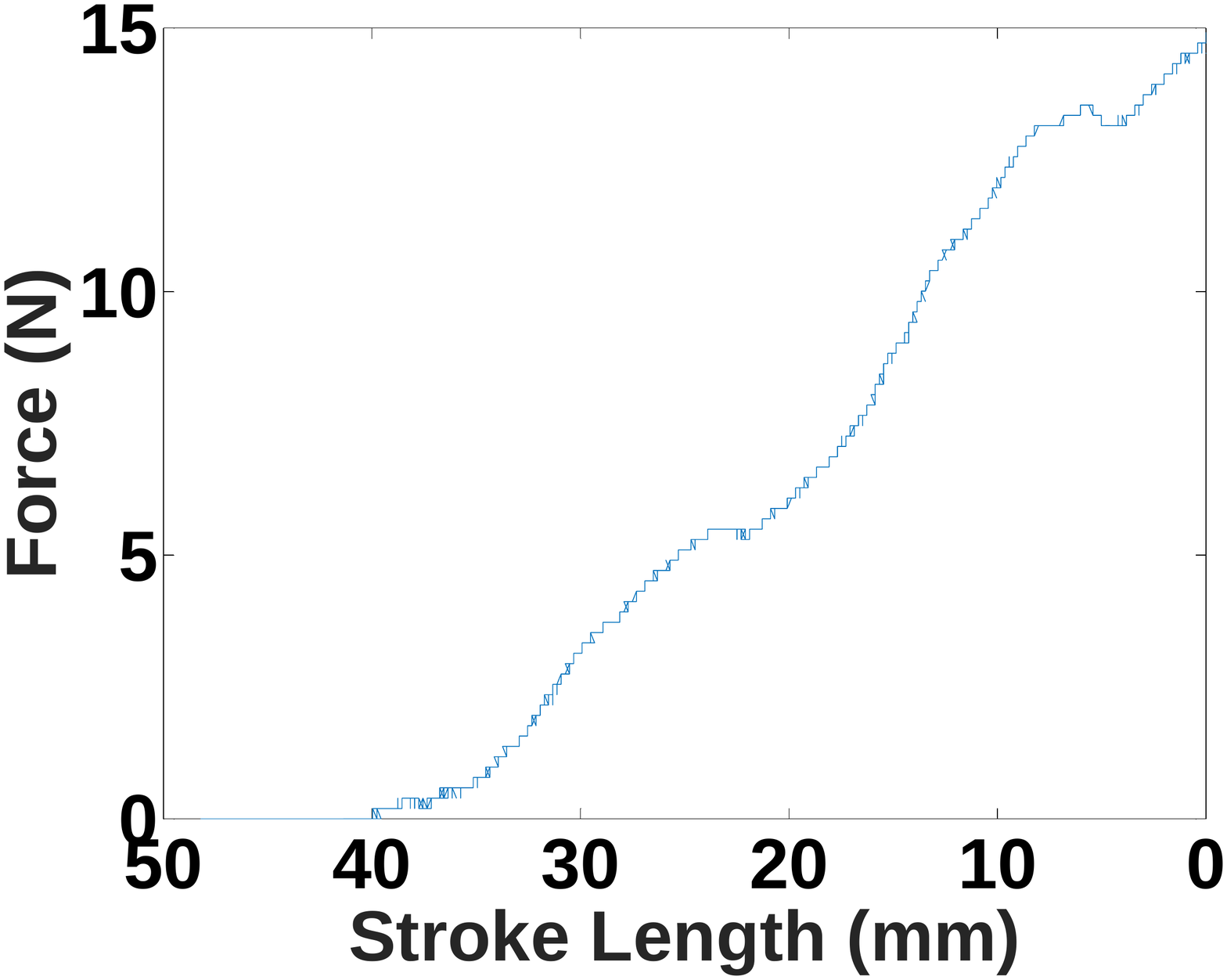} &
\includegraphics[width=0.35\linewidth]{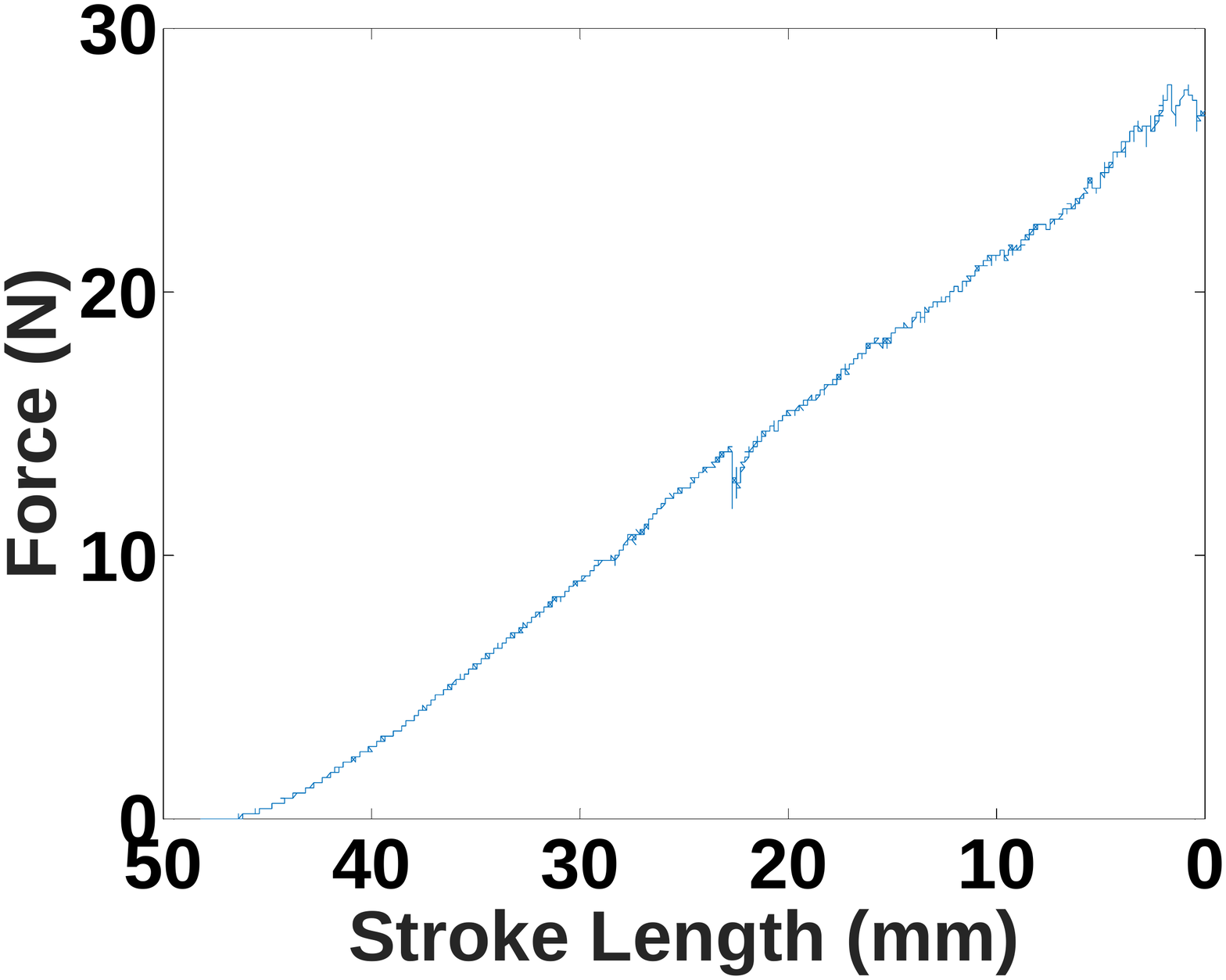} &
\includegraphics[width=0.35\linewidth]{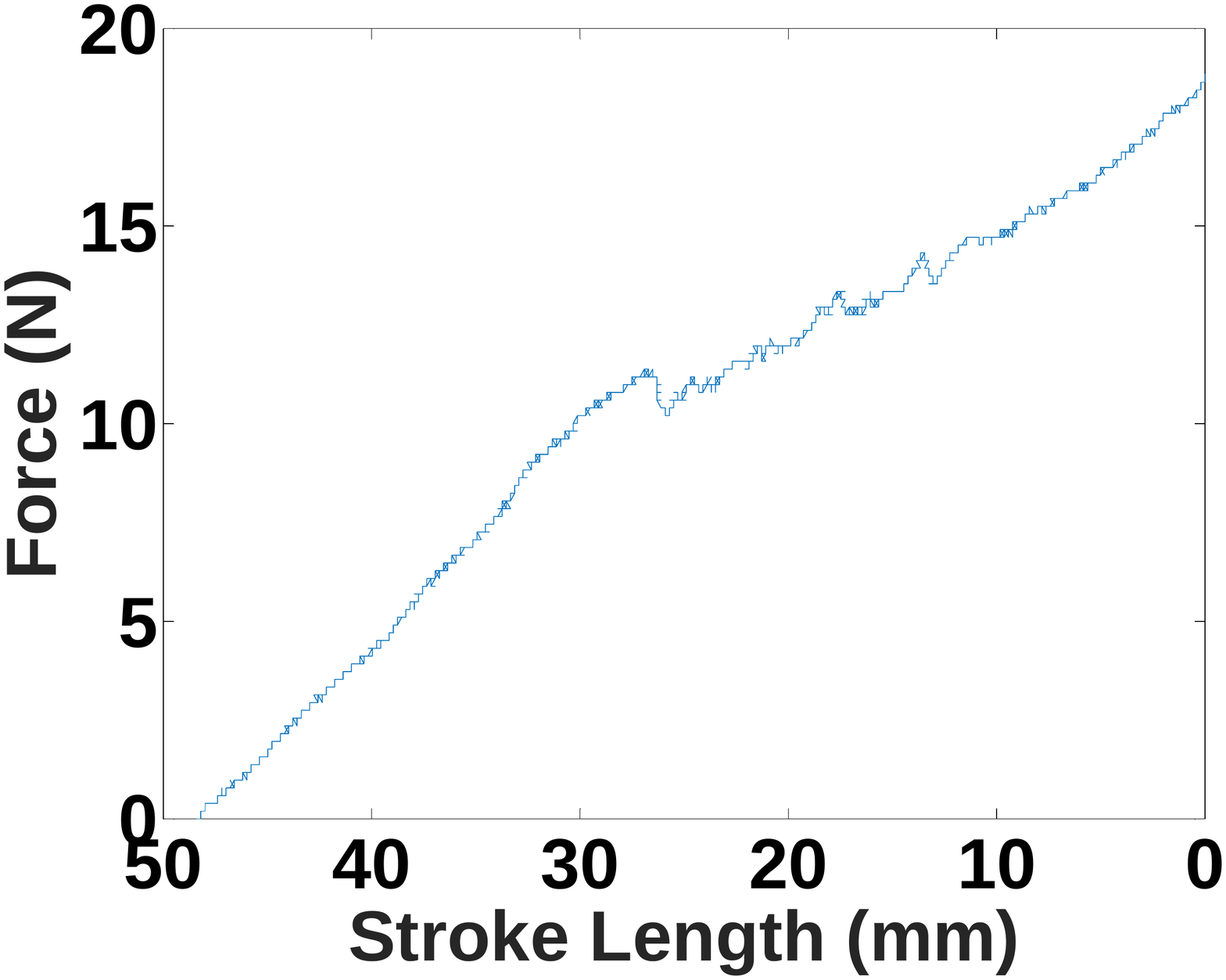} \\
\parbox{45mm}{\footnotesize Subject \#1. Functional extension was achieved. Spasticity level: 2.}&
\parbox{45mm}{\footnotesize Subject \#2. Functional extension was achieved. Spasticity level: 1.}&
\parbox{45mm}{\footnotesize Subject \#3. Functional extension was achieved. Spasticity level: 1 (except for index finger, rated at 2).}\\[3mm]
\includegraphics[width=0.35\linewidth]{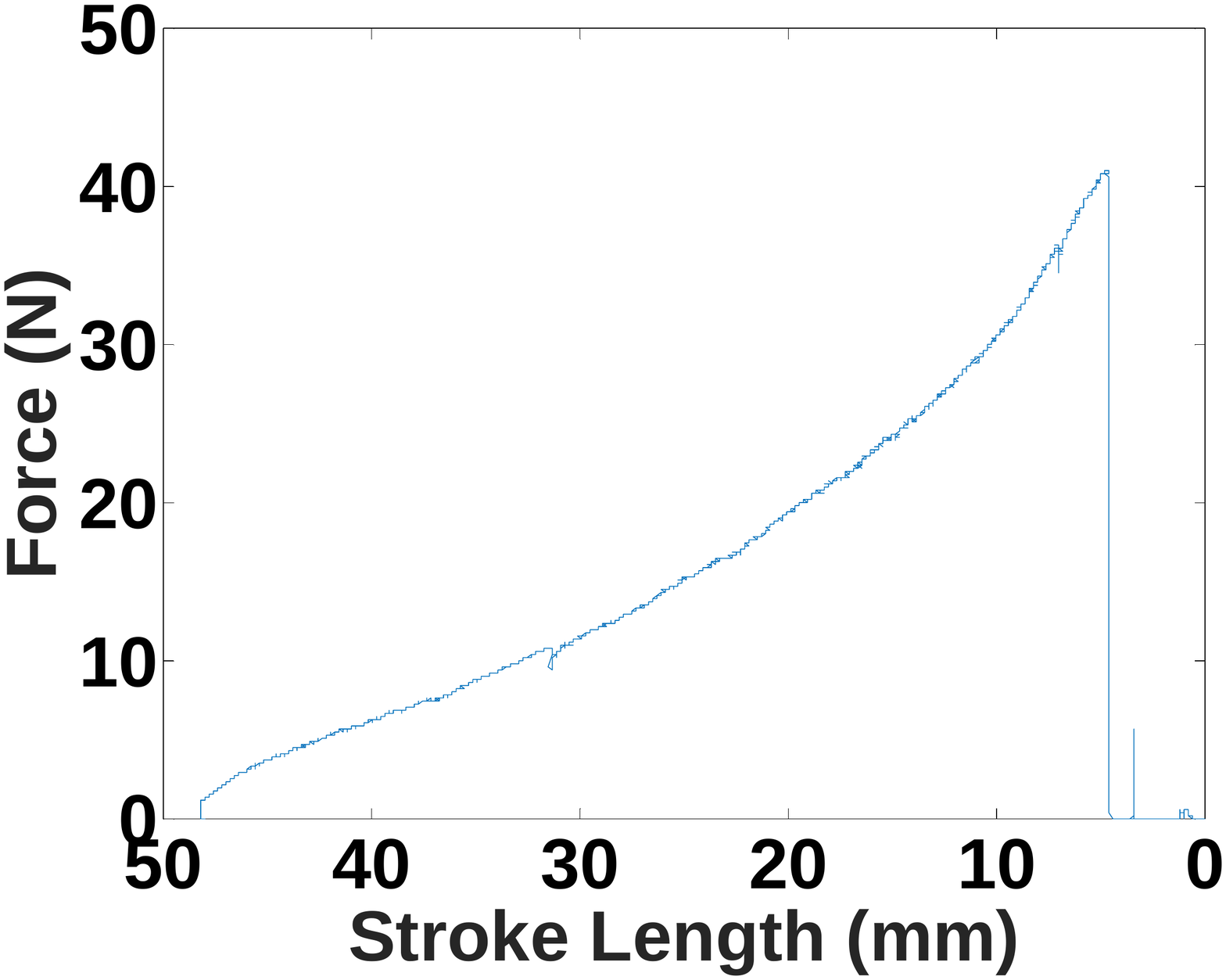} &
\includegraphics[width=0.35\linewidth]{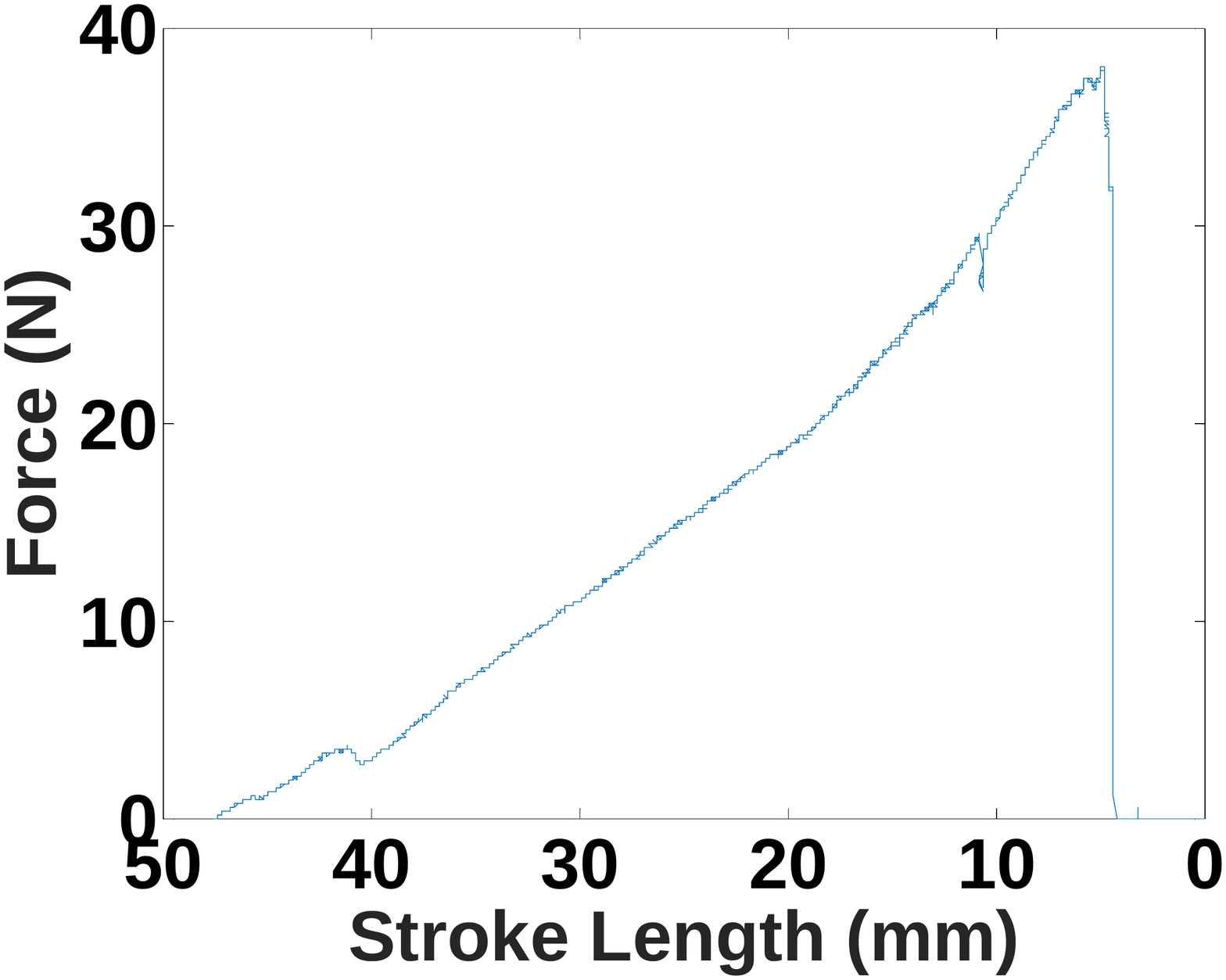} &
\\
\parbox{45mm}{\footnotesize Subject \#4. Functional extension was not achieved. Spasticity level: 3.}&
\parbox{45mm}{\footnotesize Subject \#5. Functional extension was achieved. Spasticity level: 2.}&
\\
\end{tabular}
\caption{Characterization of hand extension trials: force vs. position
  data. Each plot shows, for one trial, the relationship between the
  measured force in the actuated tendon and the linear position of the
  actuator. Note that each trial begins with the actuator fully
  elongated (50 mm actuator position) and slack tendon network (0N
  force). As the actuator retracts (left-to-right movement on the
  plots), we measure the force applied to the tendon network. If the
  force exceeds the maximum load supported by the magnet, the
  mechanism disengages producing a sudden drop in the force
  profile. This plot shows one representative trial from each of the 5
  subjects tested using this movement pattern. For each trial, we also
  indicate whether functional extension (defined as sufficient hand
  opening to grasp an object of approximately 55mm in diameter) was
  achieved before the mechanism achieved maximum retraction or
  disengaged.}
\label{fig:results}
\end{figure*}

One of the main objectives of our set of experiments was to determine
the actuation forces needed to achieve functional hand extension in
stroke patients exhibiting various levels of
spasticity. Fig.~\ref{fig:results} summarizes our results measuring
actuation forces during trials with five stroke patients using tendon
configuration 1 (full extension). We present one representative trial
per subject; surprisingly, we found very little variation between
trials were the subject was asked to relax and trials where the
subject was asked to actively assist the device or to attempt a
grasp. Throughout the trials, we recorded the applied force levels as
a function of the position of the actuator. Hand opening in response
to the device was observed by the experimenter and rated as functional
(sufficient to grasp an object of approximately 55mm in diameter) or
not; however, quantitative data for joint angles was not recorded.

The assistive hand device was able to achieve functional hand
extension for 4/5 patients. In 2/5 cases the force level led to the
breakaway mechanism disengaging; however, in one of those cases this
occurred after functional hand extension was achieved. Overall, we
were able to achieve functional hand extension for all patients with
MAS spasticity levels of 1 and 2. Breakaway occurred after achieving
functional extension for one patient with MAS spasticity at level 2,
and without achieving functional extension for one patient with MAS
spasticity at level 3. In all cases, the subject was able to complete
an enveloping grasp of the target object, as described in the previous
section.

The maximum level of recorded force varied between subjects: between
15-20 N for Subjects 1 and 3, between 25-30 N for Subject 2, and
exceeding 35 N (and thus leading to breakaway) for Subjects 4 and
5. These results suggest that an exotendon assistive glove with a
single actuator able to apply up to 40 N to a tendon network similar
to the one used here will succeed in generating functional hand
extension for most patients with spasticity levels 1 and 2, but will
not be strong enough to be used by patients with spasticity level
3.

\begin{figure}[t]
\setlength{\tabcolsep}{-2mm}
\centering
\begin{tabular}{cc}
\includegraphics[width=0.55\linewidth]{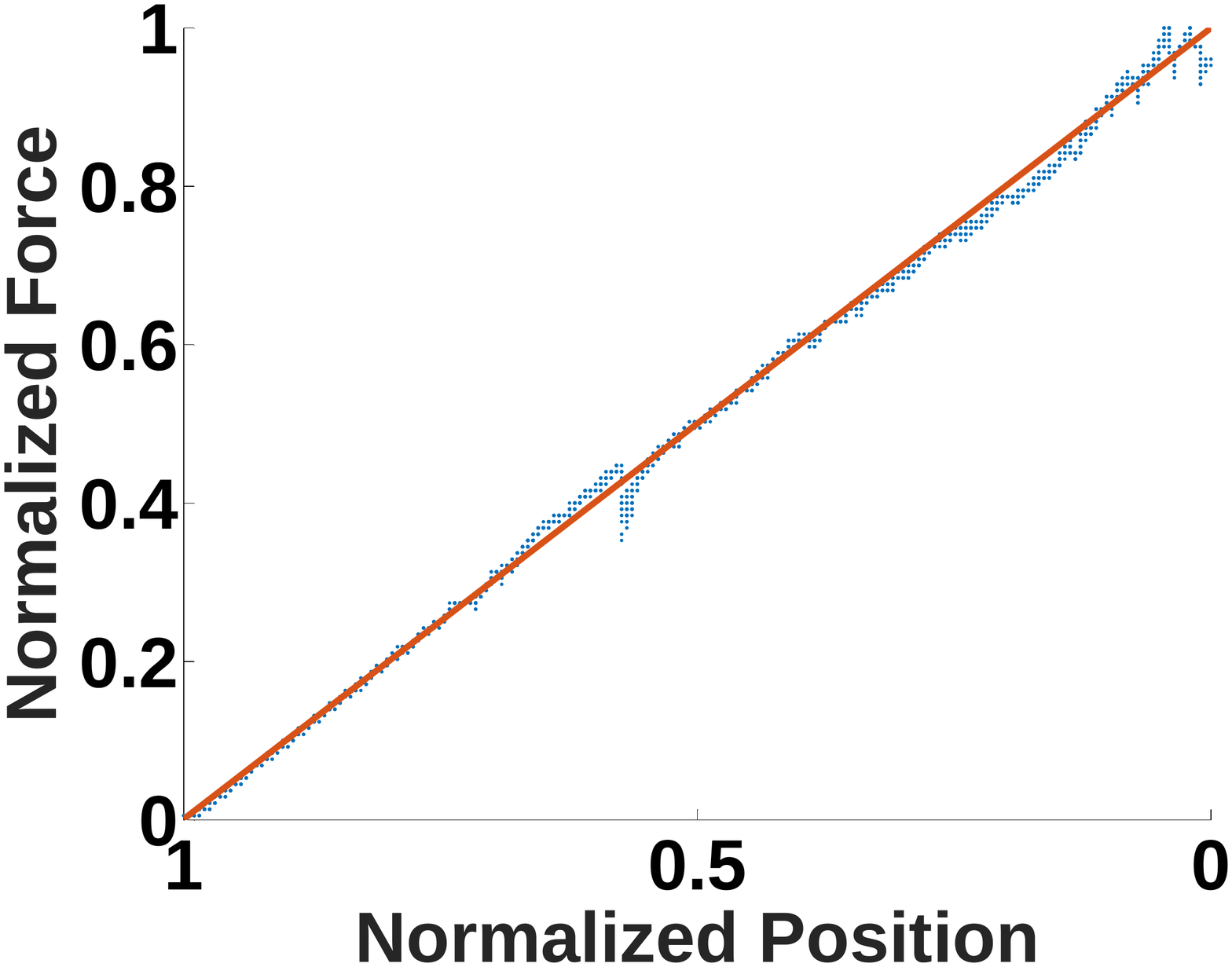}&
\includegraphics[width=0.55\linewidth]{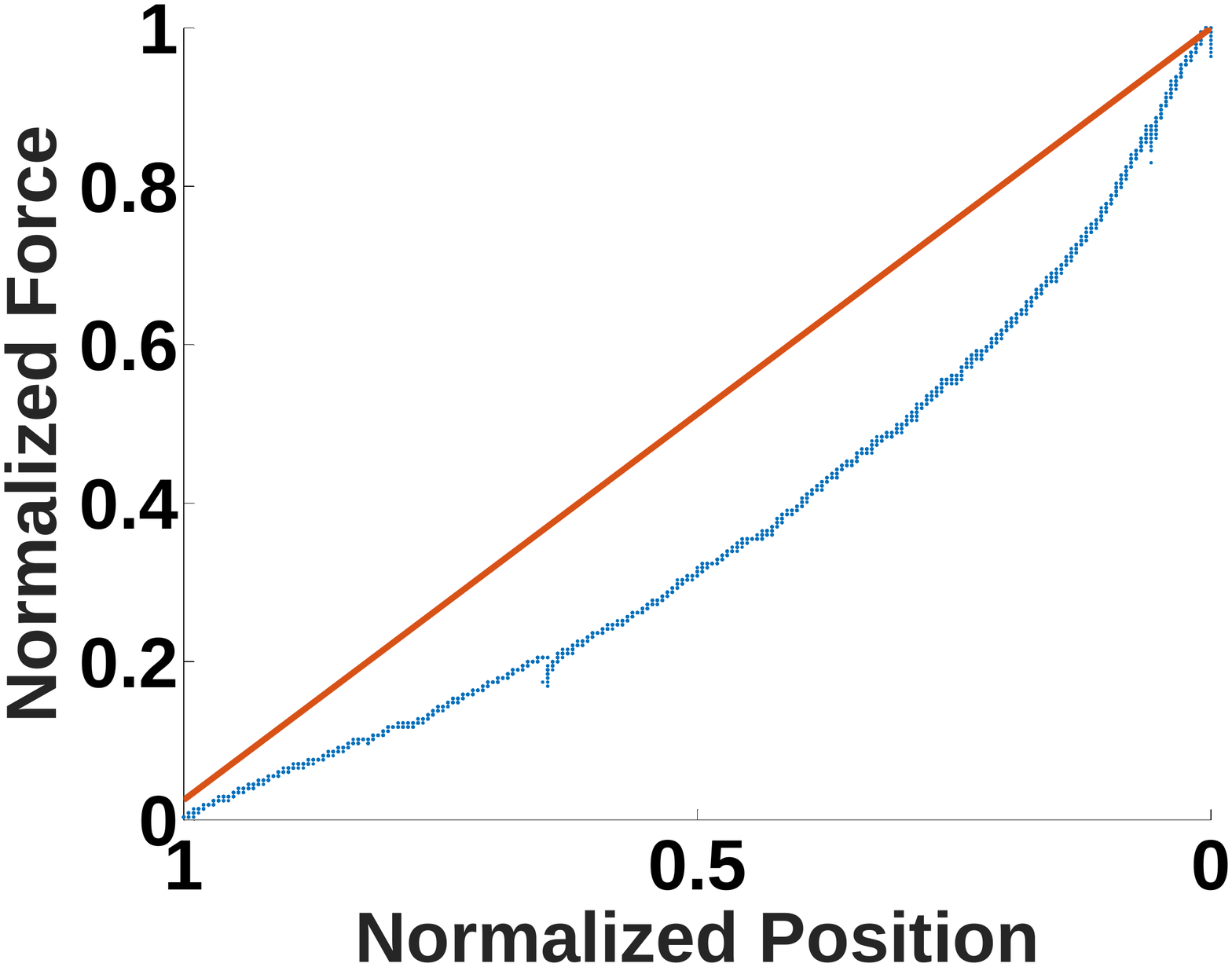}
\end{tabular}
\caption{Normalized force vs. position data (blue) and superimposed
  linear fit (red) for trials exhibiting the highest (left, 1.00) and
  lowest (right, 0.97) correlation coefficient between these two
  variables.}
\label{fig:linear}
\end{figure}

An interesting finding concerns the observed relationship between
force and position, used here as a proxy for hand pose. In all
observed cases, the relationship was highly linear. To quantify this
phenomenon, we first normalized the data as follows. First, as each
trial generally begins with a small amount of slack in the tendon that
is picked up as the actuator retracts, we removed all data points
until the force first first reached a threshold of 3N. For cases where
the trial ended by a disconnect of the breakaway mechanism, we also
removed data points starting at the breakaway moment (observed as a
sudden drop in force levels all the way to 0). Finally, we normalized
remaining force and position values by dividing with the maximum
observed value, and measured the correlation coefficient between the
resulting data series.

For all the trials shown in Fig.~\ref{fig:results}, we obtained
correlation coefficients ranging between 0.97 (Subject \#2) and 1.00
(Subject \#4). Fig.~\ref{fig:linear} shows the normalized data, along
with the linear fit, for the trials with the highest and lowest
correlation coefficient. These results suggest that the spastic muscle
does not oppose movement with a fixed force level. Rather, it behaves
in spring-like fashion, with resistance increasing linearly along with
elongation.

We also carried out experiments with two stroke patients using tendon
configuration 2 (MCP flexion / IP extension). Force vs. position data
for these experiments is less relevant, since, in these cases the
assistive device does not need to overcome involuntary forces in the
opposite direction. Rather, we were interested in the functional
aspect: does the assistive device in this configuration enable stable
fingertip grasping. This was quantified as the ability to hold the
object using such a grasp without external support while the assistive
device was engaged and applying tendon forces. One of the subjects
displayed this ability without the use of an assistive device;
however, both subjects were able to perform this task with the use of
the device.

Throughout the experiments, none of the subjects reported any pain or
discomfort from using the device. However, subjective feedback
repeatedly included suggestions to make doning the glove component of
the device easier. We plan to take more steps in this direction for
further iterations.

\section{DISCUSSION AND CONCLUSIONS}

Based on our results, we draw the following conclusions, which we
believe can be informative for research and development for next
generation wearable assistive devices for the hand:
\begin{itemize}
\item A single linear actuator applying a total force below 40 N can
  overcome hand spasticity and produced desired movement patterns for
  stroke survivors with spasticity levels 1 and 2. This suggests that
  lightweight wearable devices (we used an actuator with a total
  weight of 40 g) can be effective, from a force generation
  perspective, for a significant range of the population affected by
  hand impairments as a result of stroke.
\item Multiple functional whole-hand movement patterns can be produced
  using a single actuator for each. In particular, we demonstrated two
  such patterns: full extension (which combines with voluntary flexion
  to produce enveloping grasps) and MCP flexion / IP extension (which
  produces fingertip grasps). These results suggest that a single
  device with a small number of actuators can combine multiple such
  patterns, producing a wider range of manipulation capabilities.
\item Our results also suggest that spastic muscles oppose movement in a
  spring-like fashion, with forces increasing linearly with
  elongation. In turn, this suggests that selection of actuators (and
  implicitly force levels) for assistive devices must take into
  consideration how applied forces will vary throughout the expected
  range of motion.
\end{itemize}

Overall, by illustrating the fact that multiple movement patterns are
possible using few and portable actuators, this study provides data to
support the feasibility of building assistive devices for the hand
that will be both wearable and dexterous. Such devices could help with
the manipulation component of activities of daily living, enable
functional training on real-world manipulation tasks, extend training
beyond the small number of sessions performed in a clinical setting,
and provide distributed training during the course of daily activities
rather than block training in designated therapeutic sessions.

Our current study also revealed a number of limitations. While hand
movement was observed by the experimenter and functionally tested by
the ability to execute a grasp (enveloping or fingertip), digit
movement was not measured or recorded. As such, based on this data, we
can not study the relationship between joint angles and actuator
position or force. When performing grasps, we did not measure the
resultant forces applied to the object or numerically quantify the
stability of the resulting grasp. Finally, the study was exploratory
in nature, covering stroke patients with multiple levels of spasticity
and hand impairment types, but without a sufficient population to
examine the statistical significance of the findings. We plan to
address these limitations in future work.

\bibliographystyle{IEEEtran}
\bibliography{bib/orthosis,bib/grasping,bib/thesis}

\end{document}